\documentclass[journal,twoside,web]{ieeecolor}
\usepackage{tmech}
\usepackage{amsmath}
\usepackage{amssymb}
\usepackage{graphicx}
\usepackage[ruled,linesnumbered]{algorithm2e}
\usepackage{booktabs}
\usepackage{cuted}
\usepackage{balance}
\usepackage[hidelinks=true,bookmarks=FWLse]{hyperref}
\usepackage{lineno}
\usepackage{amsopn}
\usepackage{comment}
\usepackage{color}
\usepackage{cite}
\usepackage{algorithmic}
\usepackage{xcolor}
\usepackage{nomencl}
\let\labelindent\relax
\usepackage{enumitem}
\usepackage{url,subfigure,tabulary}
\newlist{abbrv}{itemize}{1}
\setlist[abbrv,1]{label=,labelwidth=1in,align=parleft,itemsep=0.2\baselineskip,leftmargin=!}

\SetKwRepeat{Do}{do}{while}

\newcommand{\egi}{\textit{e.g.}}

\newcolumntype{L}[1]{>{\raggedright\arraybackslash}p{#1}}
\newcolumntype{C}[1]{>{\centering\arraybackslash}p{#1}}
\newcolumntype{R}[1]{>{\raggedleft\arraybackslash}p{#1}}

\def\BibTeX{{\rm B\kern-.05em{\sc i\kern-.025em b}\kern-.08em
    T\kern-.1667em\lower.7ex\hbox{E}\kern-.125emX}}

\begin{document}
\title{Learning Collision-Free Space Detection from Stereo Images: 
	Homography Matrix Brings\\Better Data Augmentation }
\author{Rui Fan, \IEEEmembership{Member, IEEE}, Hengli Wang, \IEEEmembership{Graduate Student Member, IEEE}, Peide Cai,\\ Jin Wu, \IEEEmembership{Member, IEEE}, 
	Mohammud Junaid Bocus, Lei Qiao, Ming Liu, \IEEEmembership{Senior Member, IEEE}
\vspace{-2.5em}
\thanks{This work was supported by the National Natural Science Foundation of China, under grant No. U1713211, Collaborative Research Fund by Research Grants Council Hong Kong, under Project No. C4063-18G, and HKUST-SJTU Joint Research Collaboration Fund, under project SJTU20EG03, awarded to Prof. Ming Liu.
}
\thanks{R. Fan is with the Department of Computer Science and Engineering, as well as the Department of Ophthalmology, the University of California San Diego, La Jolla, CA 92093, United States (e-mail: rui.fan@ieee.org).}
\thanks{H. Wang, P. Cai, J. Wu and M. Liu are with the Department of Electronic and Computer Engineering, the Hong Kong University of Science and Technology, Hong Kong SAR, China (e-mail: \{hwangdf, pcaiaa, jwucp, eelium\}@ust.hk).}
\thanks{Mohammud Junaid Bocus is with the Department of Electrical and Electronic Engineering, the University of Bristol, BS8 1UB, the United Kingdom (e-mail: junaid.bocus@bristol.ac.uk).}
\thanks{L. Qiao is with the State Key Laboratory of Ocean Engineering and the School of Naval Architecture, Ocean and Civil Engineering, Shanghai Jiao Tong University, Shanghai, 200240, China (e-mail: qiaolei@sjtu.edu.cn).}
\thanks{R. Fan and H. Wang contributed equally to this work.}
}

\maketitle

\begin{abstract}
Collision-free space detection is a critical component of autonomous vehicle perception. The state-of-the-art algorithms are typically based on supervised deep learning. Their performance is dependent on the quality and amount of labeled training data. It remains an open challenge to train deep convolutional neural networks (DCNNs) using only a small quantity of training samples. Therefore, in this paper, we mainly explore an effective training data augmentation approach that can be employed to improve the overall DCNN performance, when additional images captured from different views are available. 
Due to the fact that the pixels in collision-free space (generally regarded as a planar surface) between two images, captured from different views, can be associated using a homography matrix, the target image can be transformed into the reference view. This provides a simple but effective way to generate  training data from additional multi-view images. Extensive experimental results, conducted with six state-of-the-art semantic segmentation DCNNs on three datasets, validate the effectiveness of the proposed method for enhancing collision-free space detection performance. When validated on the KITTI road benchmark, our approach provides the best results, compared with other state-of-the-art stereo vision-based collision-free space detection approaches.
\end{abstract}


\section*{List of Symbols}
\begin{abbrv}
	\item[$r,t$] pinhole cameras
	\item[$d$] disparity
	\item[$f$] camera focal length
	\item[$u$] horizontal coordinate of $\boldsymbol{p}$
	\item[$v$] vertical coordinate of $\boldsymbol{p}$
	\item[$o_u$] horizontal coordinate of $\boldsymbol{p}_o$
	\item[$o_v$] vertical coordinate of $\boldsymbol{p}_o$
	\item[$z$] depth from camera to $\boldsymbol{P}$
	\item[$n_{x,y,z}$] $x$, $y$ and $z$ coordinates of $\boldsymbol{n}$ 
	
	\item[$\Phi$] stereo rig roll angle
	\item[$\varkappa,\kappa$] road disparity projection model coefficients
	\item[$p_0-p_5, \Delta$] constants for $\Phi$ estimation
	\item[$c$] constant for $\varkappa$ and $\kappa$ estimation
	\item[$w$] image rotation function
	\item[$m$] disparity pixel number
	
	%	\item[$n_{tp}$] true positive pixel number
	%	\item[$n_{tn}$] true negative pixel number
	%	\item[$n_{fp}$] false positive pixel number
	%	\item[$n_{fn}$] false negative pixel number
	
	\item[$E$] energy for $\Phi$, $\varkappa$ and $\kappa$ estimation
	\item[$D$] distance between $r$ and the planar surface
	\item[$W$] image width
	
	\item[$T_c$] stereo rig baseline
	\item[$I$] driving scene image
	\item[$\boldsymbol{p}$] 2D image pixel
	\item[$\boldsymbol{p}_o$] principal point
	\item[$\tilde{\boldsymbol{p}}$] homogeneous coordinates of $\boldsymbol{p}$
	\item[$\boldsymbol{t}$] translation vector
	\item[$\boldsymbol{n}$] normal vector of the planar surface
	\item[$\boldsymbol{I}$] identity matrix
	\item[$\boldsymbol{P}$] 3D point in the world coordinate system
	\item[$\boldsymbol{R}_{tr}$] rotation matrix
	\item[$\boldsymbol{H}_{tr}$] homography matrix
	\item[$\boldsymbol{K}$] camera intrinsic matrix
\end{abbrv}

\begin{figure*}[!t]
	\centering
	\includegraphics[width=0.99\textwidth]{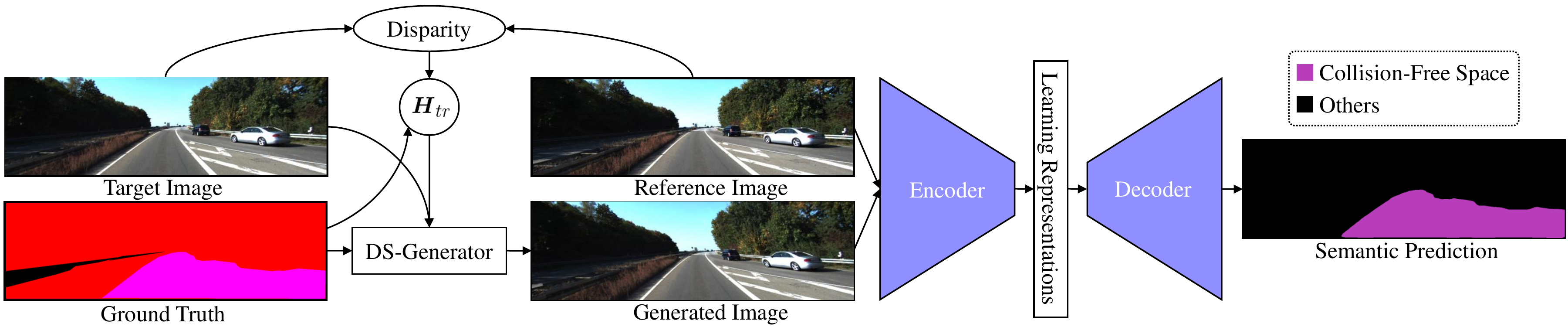}
	\caption{Block diagram of our proposed collision-free space detection approach.}
	\label{fig.blok_digram}
\end{figure*}

\section{Introduction}

\IEEEPARstart{T}HE paradigm in the automotive industry has shifted from high-performance cars to comfortable and safe cars in the past decade \cite{pieri2018consumer}. This paradigm shift has accelerated the development of autonomous driving technologies, such as the internet of vehicles (IoV) \cite{zhang2019mobile} and advanced driver assistance systems (ADAS). In recent years, industry titans, such as Waymo, BMW, Tesla and Volvo, have been competing with each other to commercialize autonomous vehicles \cite{fan2020book}. However, a number of 
accidents occurred during experiments recently, and this has cast doubt on whether the autonomous driving technology is safe enough for deployment \cite{self-driving-car-news1}. In this regard, the self-driving industry is now becoming more realistic. Many of them believe that the current research and development of autonomous driving technologies should still focus on the ADAS \cite{nagai2014research,biever2020automated}.

Visual environment perception is a key component of the ADAS \cite{fan2020book}. Its tasks include \cite{grigorescu2020survey}: a) 3D information acquisition; b) object detection/recognition; and c) semantic segmentation. Collision-free space detection, also referred to as occupancy grid mapping or drivable area detection, is an important task in visual environment perception \cite{sless2019road}. Collision-free space detection approaches generally classify each pixel in the image as positive (drivable) or negative (undrivable) \cite{wang2021dynamic}. Such classification results are then used by other autonomous car modules, \egi, trajectory prediction \cite{thiede2019analyzing}, lane departure warning \cite{fan2018lane_detection}, and obstacle avoidance \cite{pouyanfar2019roads}, to ensure that the autonomous car can safely navigate in complex environments.

Recent deep convolutional neural network (DCNN)-based collision-free space detection approaches perform incredibly well \cite{fan2020sne, wang2020applying}. However, the quality and amount of training samples can greatly affect the performance of these DCNNs. In this regard, training data augmentation is generally performed to increase the diversity of the available data, without actually collecting new data. The most common way of training data augmentation is to apply different types of image transformation operations, such as reflections, rotations and translations, to the existing data. Fortunately, for a multi-camera system, such as a stereo rig, multi-view images are available. However, the aforementioned image transformation operations do not consider the relationship among images captured at different view points. Therefore, jointly exploring effective training data augmentation approaches and leveraging the relationship among multi-view images, especially for stereo images, has become a popular area of research that requires more attention.

The collision-free space can be considered as a planar surface. Since the 3D points on the same planar surface between two images captured from different views can be linked by a homography matrix \cite{hartley2003multiple}, the target image can be transformed into its reference view \cite{fan2018road}. Hence in this paper, we propose an effective driving scene generator (DS-Generator), which can produce additional RGB images for training data augmentation. The block diagram of our proposed collision-free space detection approach is shown in Fig. \ref{fig.blok_digram}. The 3D points on the collision-free space between the reference and target images are first used to estimate their corresponding homography matrix. The target image and the estimated homography matrix then serve as the input to our DS-Generator, and a driving scene image can be generated. Since the generated image is in the same view of the reference image, they can use the same ground truth label. To validate the effectiveness of our DS-Generator, we train six state-of-the-art semantic segmentation DCNNs on three road segmentation datasets for collision-free space detection. Extensive experiments illustrate that our DS-Generator can effectively augment training sets and all the evaluated DCNNs achieve better results for collision-free space detection. When validated on the KITTI road benchmark\footnote{\url{www.cvlibs.net/datasets/kitti/eval_road.php}} \cite{Fritsch2013ITSC}, our approach provides the best results, compared with other state-of-the-art stereo vision-based collision-free space detection approaches.

The remainder of this paper is organized as follows: Sec. \ref{sec.related_work} provides an overview of the state-of-the-art collision-free space detection approaches. Sec. \ref{sec.algorithm_description} introduces our DS-Generator for training data augmentation. Sec. \ref{sec.experimental_results} shows the experimental results of the six state-of-the-art DCNNs and demonstrates the effectiveness of our DS-Generator for enhancing collision-free space detection. Finally, Sec. \ref{sec.conclusion_future_work} summarizes the paper.

\section{Related Work}
\label{sec.related_work}

The state-of-the-art collision-free space detection algorithms are generally grouped into two classes: a) geometry-based and b) deep learning-based. The geometry-based algorithms typically formulate collision-free space with an explicit geometry model, {\egi}, a straight line \cite{fan2019real} or a quadratic surface \cite{fan2019pothole}, and find its best coefficients using optimization approaches, such as gradient descent \cite{fan2019real} or singular value decomposition (SVD) \cite{fan2019pothole}. The collision-free space can then be detected by comparing the difference between the actual and modeled road surfaces \cite{fan2019pothole}. \cite{wedel2009b} is a typical geometry-based collision-free space detection algorithm, where the road segmentation was performed by fitting a B-spline model \cite{knott2000interpolating} to the road disparity projections on a 2D disparity histogram (referred to as \textit{v-disparity image} \cite{labayrade2002real}). Similarly, \cite{zhang2018dijkstra} considered road surface modeling as a shortest path problem and extracted the road disparity projections from the v-disparity image using Dijkstra algorithm \cite{goldberg1993heuristic}. Moreover, \cite{fan2019pothole} and \cite{fan2019tits} formulated the road disparity projection modeling into a more general way by incorporating the stereo rig roll angle into the least squares fitting process, which can produce more robust results when the stereo rig baseline is not perfectly parallel to the collision-free space \cite{fan2019tits}.

With recent advances in machine learning, collision-free space detection is regarded as a part of semantic driving scene segmentation, where DCNNs are proven to be the best solution. Since \cite{long2015fully} introduced Fully Convolutional Network (FCN), research on semantic driving scene segmentation has experienced a major boost. SegNet \cite{badrinarayanan2017segnet} presented the encoder-decoder architecture, which is widely utilized in current networks. The encoder network performs convolutions and max-poolings, while the decoder network uses the transferred pooling indices from the encoder to produce a sparse feature map, which is then fed to a trainable filter bank to produce a dense feature map \cite{badrinarayanan2017segnet}. Finally, a softmax classifier is used for the classification of each image pixel. U-Net \cite{ronneberger2015u} was designed based on FCN \cite{long2015fully}. It consists of a contracting path and an expansive path \cite{ronneberger2015u}. The former includes convolutions, rectified linear units, and max pooling layers, while the latter combines the feature and spatial information through a sequence of upconvolutions and concatenations with the corresponding feature map from the contracting path \cite{ronneberger2015u}. 

DeepLabv3+ \cite{chen2018encoder} was improved from DeepLabv1 \cite{chen2014semantic}, DeepLabv2 \cite{chen2017deeplab} and DeepLabv3 \cite{chen2017rethinking}. It was designed to combine the advantages of both the spatial pyramid pooling (SPP) module and the encoder-decoder architecture. It applies the depthwise separable convolution to both atrous SPP (ASPP) and the decoder module, which makes its encoder-decoder module much faster and more robust \cite{chen2018encoder}.
In \cite{chen2017deeplab}, ASPP was proposed to concatenate multiple atrous-convolved features into a final feature map. However, the feature resolution is not dense enough for  semantic driving scene segmentation. DenseASPP \cite{yang2018denseaspp} was proposed to solve this problem, by connecting a set of atrous convolutional layers (ACLs) in a dense way. The ACLs in DenseASPP are organized in a cascade fashion, where the dilation rate increases layer by layer \cite{yang2018denseaspp}. Then, DenseASPP concatenates the output of each atrous layer with the input feature map and all the outputs from lower layers.
The final output of DenseASPP is a feature map generated by multi-scale atrous convolutions \cite{yang2018denseaspp}.
For recent approaches with encoder-decoder architectures, the last layer of the decoder is typically a bilinear upsampling procedure for final pixel-wise prediction recovery. 

However, the simple bilinear upsampling has limited ability to accurately recover the pixel-wise prediction, because it does not take the correlation among the prediction of each pixel into account \cite{tian2019decoders}. Data-dependent upsampling (DUpsampling) \cite{tian2019decoders} was designed to solve this problem, by exploiting the redundancy in the label space of semantic image segmentation and recovering the pixel-wise prediction from low-resolution outputs of DCNNs. Due to the effectiveness of DUpsampling, the encoder can avoid the excessive reduction of its overall strides and this can in turn reduce the consumption of computation and memory resources dramatically \cite{tian2019decoders}. 

Different from the aforementioned DCNNs, Gated-SCNN (GSCNN) \cite{takikawa2019gated} utilizes a novel two-branch architecture, which consists of a shape branch and a regular branch. Specifically, the regular branch can be any backbone architecture, and the shape branch processes the shape information in parallel to the regular branch through a set of residual blocks and gated convolutional layers (GCL). Then, GSCNN uses the higher-level activations in the regular branch to effectively help the shape branch only focus on the relevant boundary information \cite{takikawa2019gated}. Finally, GSCNN employs an ASPP to combine the information from the two streams in a multi-scale fashion.

\begin{figure*}[!t]
	\begin{center}
		\centering
		\includegraphics[width=0.99\textwidth]{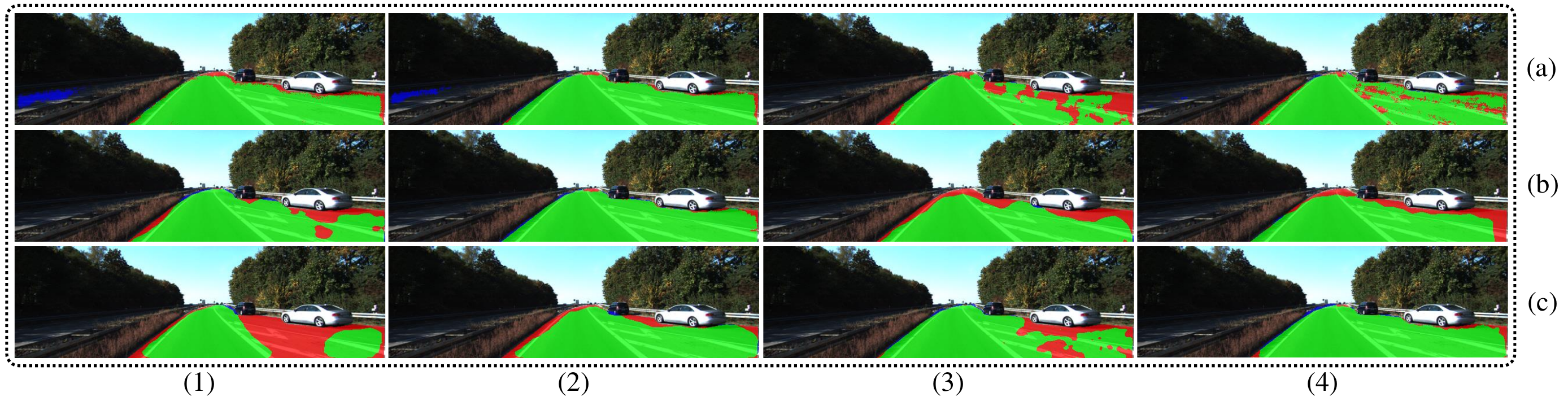}
		\centering
		\caption{Examples of the experimental results on the KITTI road dataset \cite{Fritsch2013ITSC}: columns (1)-(2)  on rows (a)-(c)  show the experimental results of (a) SegNet \cite{badrinarayanan2017segnet}, (b) DeepLabv3+ \cite{chen2018encoder} and (c) DUpsampling \cite{tian2019decoders}, trained on the original and augmented training sets, respectively; columns (3)-(4) on rows (a)-(c)  show the experimental results of (a) U-Net \cite{ronneberger2015u}, (b) DenseASPP \cite{yang2018denseaspp} and (c) GSCNN \cite{takikawa2019gated}, trained on the original and augmented training sets, respectively. The true positive, false negative and false positive pixels are shown in green, red and blue, respectively.}
		\label{fig.kitti_valid}
	\end{center}
\end{figure*}
\begin{table*}[t]
	\centering
	\caption{Performance Comparison ($\%$) Among Different DCNNs Trained on the original and augmented KITTI Road Datasets \cite{Fritsch2013ITSC}. Best Results of Each Network are shown in bold type.}
	\label{tab.kitti}
	\begin{tabular}{L{2.6cm}C{1.5cm}C{1.5cm}C{1.5cm}C{1.5cm}C{1.5cm}}
		\toprule
		Network &
		Accuracy &
		Precision &
		Recall &
		F-Score &
		IoU \\
		\midrule
		SegNet \cite{badrinarayanan2017segnet} & 93.8 & 77.6 & 85.3 & 81.2 & 68.4 \\
		HA-SegNet & \textbf{95.6} & \textbf{85.1} & \textbf{87.3} & \textbf{86.2} & \textbf{75.7} \\ \midrule
		UNet \cite{ronneberger2015u} & 95.7 & \textbf{89.6} & 82.4 & 85.9 & 75.2 \\
		HA-U-Net & \textbf{96.5} & 84.4 & \textbf{95.4} & \textbf{89.5} & \textbf{81.1} \\ \midrule
		DeepLabv3+ \cite{chen2018encoder} & 98.0 & 91.5 & \textbf{96.4} & 93.9 & 88.5 \\
		HA-DeepLabv3+ & \textbf{98.6} & \textbf{97.2} & 93.9 & \textbf{95.5} & \textbf{91.4} \\ \midrule
		DenseASPP \cite{yang2018denseaspp} & 97.3 & 90.8 & 92.0 & 91.4 & 84.1 \\
		HA-DenseASPP & \textbf{98.5} & \textbf{93.9} & \textbf{96.4} & \textbf{95.1} & \textbf{90.7} \\ \midrule
		DUpsampling \cite{tian2019decoders} & 94.7 & 82.5 & 83.8 & 83.1 & 71.2 \\
		HA-DUpsampling & \textbf{96.2} & \textbf{90.2} & \textbf{85.2} & \textbf{87.7} & \textbf{78.0} \\ \midrule
		GSCNN \cite{takikawa2019gated} & 94.8 & 84.1 & 82.4 & 83.2 & 71.3 \\
		HA-GSCNN & \textbf{95.4} & \textbf{87.1} & \textbf{83.2} & \textbf{85.1} & \textbf{74.1} \\
		\bottomrule
	\end{tabular}
\end{table*}

\section{Methodology}
\label{sec.algorithm_description}

We have two pinhole cameras $r$ and $t$,\footnote{$r$ and $t$ refer to ``reference'' and ``target'', respectively.} looking at a 3D  point $\boldsymbol{P}_i$ on a planar surface in the world coordinate system (WCS). The image pixel $^r{\boldsymbol{p}}_i=({^r}u_i;{^r}v_i)$ of $\boldsymbol{P}_i$ captured by $r$ and the image pixel ${^t}{\boldsymbol{p}}_i=({^t}u_i;{^t}v_i)$ of  $\boldsymbol{P}_i$  captured by $t$ can be linked using \cite{fan2018road}
\begin{equation}
	^t\tilde{\boldsymbol{p}}_i= \boldsymbol{H}_{tr} {^r}\tilde{\boldsymbol{p}}_i,
\end{equation}
where ${^{r,t}}\tilde{\boldsymbol{p}}$ is the homogeneous coordinates of ${^{r,t}}{\boldsymbol{p}}$, and the expression of the homograph matrix $\boldsymbol{H}_{tr}$ is \cite{hartley2003multiple}:
\begin{equation}
	\boldsymbol{H}_{tr}=\frac{{^r}z_i}{^tz_i}\boldsymbol{K}_t\cdot\Big(\boldsymbol{R}_{tr}-\frac{\boldsymbol{t}_{tr}\boldsymbol{n}^\top}{D}\Big)\cdot\boldsymbol{K}^{-1}_{r},
	\label{eq.H1}
\end{equation}
where
$^rz_i$ and $^tz_i$ are the $z$ coordinates of $\boldsymbol{P}_i$ in the $r$ and $t$ camera coordinates systems (CCSs), respectively; $\boldsymbol{R}_{tr}$ is the rotation matrix by which $r$ is rotated with respect to $t$; $\boldsymbol{t}_{tr}$ is the translation vector from $r$ to $t$; $\boldsymbol{K}_r$ and $\boldsymbol{K}_t$ are the intrinsic matrices of $r$ and $t$, respectively; $\boldsymbol{n}=(n_x;n_y;n_z)$ is the normal vector of the collision-free space; and $D$ is the distance between $r$ and the collision-free space. For a stereo rig, ${^r}z_i={^t}z_i$, $\boldsymbol{R}_{tr}$, $\boldsymbol{t}_{tr}$, $\boldsymbol{K}_r$ and $\boldsymbol{K}_t$ can be obtained from stereo rig calibration, $\boldsymbol{R}_{tr}=\boldsymbol{I}$, and $\boldsymbol{t}_{tr}=(Tc; 0; 0)$, where $Tc$ is the stereo rig baseline, 
\begin{equation}
	\boldsymbol{K}_r=\boldsymbol{K}_t=\begin{bmatrix}
		f & 0 & o_u\\ 0 & f & o_v \\ 0 & 0 & 1
	\end{bmatrix},
\end{equation}
$f$ is the camera focal length, and $\boldsymbol{p}_o=(o_u,o_v)$ is the principal point. (\ref{eq.H1}) can, therefore, be rewritten as: 
%\begin{strip}
\begin{equation}
	%\begin{split}
	\boldsymbol{H}_{tr}=
	\begin{bmatrix}
		1-\frac{Tc n_x}{D} & -\frac{Tc n_y}{D} & \frac{o_uTc n_x}{D}+\frac{o_v Tc  n_y}{D}-\frac{f Tc  n_z}{D}\\ 0 & 1 & 0 \\ 0 & 0 & 1
	\end{bmatrix}.
	%\end{split}
	\label{eq.H2}
\end{equation}
%\end{strip}
(\ref{eq.H2}) can be further written in a simplified form as follows \cite{fan2020we}:
\begin{equation}
	\begin{split}
		\boldsymbol{H}_{tr}=\varkappa
		\begin{bmatrix}
			\frac{1}{\varkappa}+\sin \Phi & -\cos \Phi &  -\kappa \\ 0 & 1/\varkappa & 0 \\ 0 & 0 & 1/\varkappa
		\end{bmatrix},
	\end{split}
	\label{eq.H3}
\end{equation}
where $\Phi$ is the stereo rig roll angle, $\varkappa$ and $\kappa$ are two road disparity projection model coefficients \cite{ozgunalp2016multiple}. They can be estimated by minimizing \cite{fan2021rethinking}: 
\begin{equation}
	E(\Phi,\varkappa,\kappa)=\sum_{i=1}^{m}\Bigg(  d_i - \varkappa\bigg(  w({^r}\boldsymbol{p}_i,\Phi) +\kappa \bigg)                   \Bigg)^2,
	\label{eq.E}
\end{equation}
where 
\begin{equation}
	w({^r}\boldsymbol{p}_i,\Phi)={^r}v_i\cos\Phi -  {^r}u_i \sin\Phi.
\end{equation}
$\min E(\Phi,\varkappa,\kappa)$ has a closed-form solution \cite{fan2020we}:
\begin{equation}
	\begin{split}
		\Phi=\arctan\Big(\frac{p_4p_1-p_3p_2+q\sqrt{\Delta}}{p_3p_0+p_5p_2-p_5p_1-p_4p_0} \Big)\ s.t. \ q\in\{-1,1\},
	\end{split}
	\label{eq.theta}
\end{equation}
\begin{equation}
	\varkappa=  \frac{1}{c}\bigg( m \sum_{i=1}^m d_i {w({^r}\boldsymbol{p}_i,\Phi)}     - \sum_{i=1}^m d_i  \sum_{i=1}^m {w({^r}\boldsymbol{p}_i,\Phi)}   \bigg),
\end{equation}

\begin{equation}
	\begin{split}
		\kappa&=\frac{1}{\varkappa c}\Bigg(\sum_{i=1}^m d_i \sum_{i=1}^m {w({^r}\boldsymbol{p}_i,\Phi)}^2 \\&- \sum_{i=1}^m w({^r}\boldsymbol{p}_i,\Phi)\sum_{i=1}^m  d_i  w({^r}\boldsymbol{p}_i,\Phi)    \Bigg),
	\end{split}
\end{equation}
where 
\begin{equation}
	c=    m  \sum_{i=1}^m {w({^r}\boldsymbol{p}_i,\Phi)}^2   - \Big(\sum_{i=1}^m {w({^r}\boldsymbol{p}_i,\Phi)}\Big)^2.
\end{equation}
The expressions of $p_0$--$p_5$ and $\Delta$ are given in \cite{fan2019tits}. $\Phi$ can be determined by separately replacing $q$ in (\ref{eq.theta}) with -1 and 1  and finding the minimum $\min E$ \cite{fan2019tits}. With the estimated $\Phi$, $\varkappa$ and $\kappa$, the target image ${^t}I$ can be used to generate an image ${^g}I$ in the reference view using: 
\begin{equation}
	{^g}I(\boldsymbol{p}_i)=\begin{cases}
		{^r}I(\boldsymbol{p}_i) \ \ \ \text{if}\ u_i-(\varkappa (  w(\boldsymbol{p}_i,\Phi) +\kappa )\leq0\\ 
		\ \ \ \ \ \ \ \ \ \ \text{or} \ u_i-(\varkappa (  w(\boldsymbol{p}_i,\Phi) +\kappa ) >W\\	
		{^t}I(\boldsymbol{p}_i-(\varkappa (  w({^r}\boldsymbol{p}_i,\Phi) +\kappa );0))\ \  \text{otherwise}\\
	\end{cases},
\end{equation}
where $\boldsymbol{p}_i$ is a 2D pixel in the generated image ${^g}I$ and $W$ is the image width. ${^r}I$ and ${^g}I$ then use the ground truth label of ${^r}I$ to train the DCNN. 
\begin{figure*}[!t]
	\begin{center}
		\centering
		\includegraphics[width=0.99\textwidth]{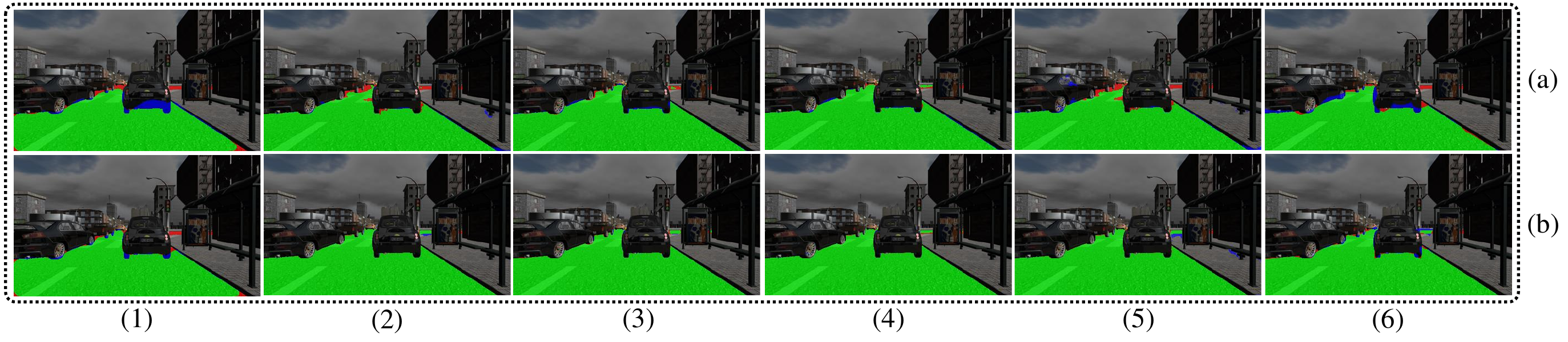}
		\centering
		\caption{Examples of the experimental results on the SYNTHIA road dataset \cite{HernandezBMVC17}, where (1) SegNet \cite{badrinarayanan2017segnet}, (2) U-Net \cite{ronneberger2015u}, (3) DeepLabv3+ \cite{chen2018encoder}, (4) DenseASPP \cite{yang2018denseaspp}, (5) DUpsampling \cite{tian2019decoders}, (6) GSCNN \cite{takikawa2019gated}, (a) trained on the original training set, and (b) trained on the augmented training set. The true positive, false negative and false positive pixels are shown in green, red and blue, respectively.}
		\label{fig.synthia}
	\end{center}
\end{figure*}
\begin{table*}[t]
	\centering
	\caption{Performance Comparison ($\%$) Among Different DCNNs Trained on the original and augmented  SYNTHIA Road Datasets \cite{HernandezBMVC17}. Best Results of Each Network are shown in bold type.}
	\label{tab.synthia}
	\begin{tabular}{L{2.6cm}C{1.5cm}C{1.5cm}C{1.5cm}C{1.5cm}C{1.5cm}}
		\toprule
		Network &
		Accuracy &
		Precision &
		Recall &
		F-Score &
		IoU \\
		\midrule
		SegNet \cite{badrinarayanan2017segnet} & 94.1 & 94.5 & 89.5 & 91.9 & 85.1 \\
		HA-SegNet & \textbf{96.3} & \textbf{95.5} & \textbf{94.2} & \textbf{94.8} & \textbf{90.2} \\ \midrule
		UNet \cite{ronneberger2015u} & 94.9 & 94.9 & 91.3 & 93.1 & 87.0 \\
		HA-U-Net & \textbf{97.1} & \textbf{95.8} & \textbf{96.1} & \textbf{95.9} & \textbf{92.2} \\ \midrule
		DeepLabv3+ \cite{chen2018encoder} & 97.2 & 95.0 & 97.4 & 96.2 & 92.7 \\
		HA-DeepLabv3+ & \textbf{98.3} & \textbf{96.8} & \textbf{98.6} & \textbf{97.7} & \textbf{95.5} \\ \midrule
		DenseASPP \cite{yang2018denseaspp} & 96.0 & 94.0 & 95.1 & 94.5 & 89.7 \\
		HA-DenseASPP & \textbf{97.7} & \textbf{95.8} & \textbf{97.8} & \textbf{96.8} & \textbf{93.8} \\ \midrule
		DUpsampling \cite{tian2019decoders} & 95.9 & 95.7 & 93.1 & 94.4 & 89.4 \\
		HA-DUpsampling & \textbf{97.4} & \textbf{95.9} & \textbf{96.9} & \textbf{96.4} & \textbf{93.0} \\ \midrule
		GSCNN \cite{takikawa2019gated} & 95.5 & \textbf{96.4} & 91.4 & 93.8 & 88.4 \\
		HA-GSCNN & \textbf{97.3} & 95.3 & \textbf{97.2} & \textbf{96.2} & \textbf{92.8} \\
		\bottomrule
	\end{tabular}
\end{table*}

\section{Experimental Results}
\label{sec.experimental_results}

\subsection{Datasets}
We conduct the experiments on three datasets:
\begin{itemize}
	\item The KITTI road dataset \cite{Fritsch2013ITSC}: this dataset provides stereo image pairs, collected in real-world environments. We split it into three sets: a) training (173 pairs of stereo images), b) validation (58 pairs of stereo images), and c) testing (58 pairs of stereo images). The disparity information is acquired by PSMNet \cite{Chang2018}.  
	\item The SYNTHIA road dataset \cite{HernandezBMVC17}: this dataset provides stereo image pairs acquired in simulation environments. We select 300 images from it and split them into three sets: training (180 pairs of stereo images), validation (60 pairs of stereo images), and testing (60 pairs of stereo images). This dataset provides the disparity ground truth. 
	\item Our SYN-Stereo road dataset: we publish a multi-view synthetic dataset, named SYN-Stereo road dataset. This dataset is created using CARLA\footnote{\url{carla.org}} simulator \cite{dosovitskiy2017carla}. We first mount a simulated stereo rig (baseline: 1.5 m) on the top of a vehicle to capture synchronized stereo images (resolution:  640$\times$480 pixels). The vehicle then navigates in different maps under different illumination and weather conditions, \egi, clear, rainy, daytime and sunset, for driving scene collection. We set random pedestrians including adults and children walking along the sidewalks. We also randomly set different types of vehicles, such as cars and motorcyclists, navigating in the scenarios at different speeds. The pedestrians and vehicles are all controlled by the CARLA simulator. We select 300 pairs of stereo images with corresponding disparity and semantic segmentation ground truth for collision-free space detection. We split them into three sets: a) training (180 pairs of stereo images), b) validation (60 pairs of stereo images), and c) testing (60 pairs of stereo images). Our dataset is publicly available at \url{sites.google.com/view/syn-stereo} for research purposes.
\end{itemize}

Please note that the training, validation, and testing sets contain data from different driving scenarios, and therefore data corresponding to a single driving scenario is only contained within one of these sets.

\begin{figure*}[!t]
	\begin{center}
		\centering
		\includegraphics[width=0.99\textwidth]{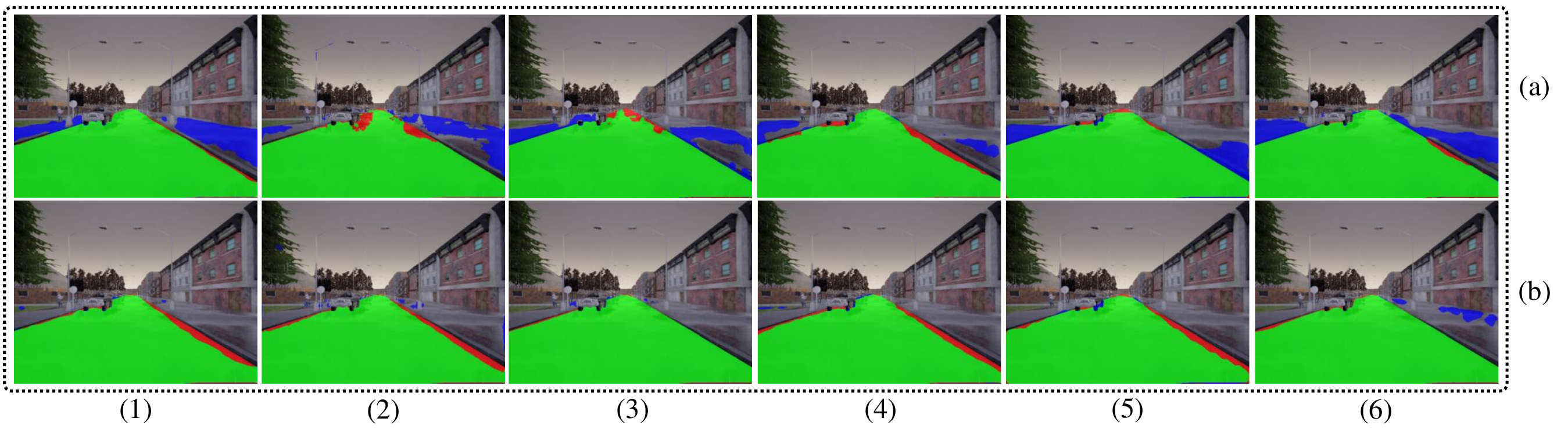}
		\centering
		\caption{Examples of the experimental results on our created SYN-Stereo road dataset, where (1) SegNet \cite{badrinarayanan2017segnet}, (2) U-Net \cite{ronneberger2015u}, (3) DeepLabv3+ \cite{chen2018encoder}, (4) DenseASPP \cite{yang2018denseaspp}, (5) DUpsampling \cite{tian2019decoders}, (6) GSCNN \cite{takikawa2019gated}, (a) trained on the original training set, and (b) trained on the augmented training set. The true positive, false negative and false positive pixels are shown in green, red and blue, respectively.}
		\label{fig.carla}
	\end{center}
\end{figure*}
\begin{table*}[t]
	\centering
	\caption{Performance Comparison ($\%$) Among Different DCNNs Trained on the original and augmented SYN-Stereo Road Datasets. Best Results of Each Network are shown in bold type.}
	\label{tab.carla}
	\begin{tabular}{L{2.6cm}C{1.5cm}C{1.5cm}C{1.5cm}C{1.5cm}C{1.5cm}}
		\toprule
		Network &
		Accuracy &
		Precision &
		Recall &
		F-Score &
		IoU \\
		\midrule
		SegNet \cite{badrinarayanan2017segnet} & 93.0 & 90.7 & 92.7 & 91.7 & 84.7 \\
		HA-SegNet & \textbf{95.6} & \textbf{96.6} & \textbf{93.0} & \textbf{94.8} & \textbf{90.1} \\ \midrule
		UNet \cite{ronneberger2015u} & 92.8 & 90.3 & 92.6 & 91.4 & 84.2 \\
		HA-U-Net & \textbf{95.4} & \textbf{95.8} & \textbf{93.4} & \textbf{94.6} & \textbf{89.7} \\ \midrule
		DeepLabv3+ \cite{chen2018encoder} & 95.3 & 95.8 & 93.1 & 94.4 & 89.4 \\
		HA-DeepLabv3+ & \textbf{97.1} & \textbf{98.2} & \textbf{95.0} & \textbf{96.6} & \textbf{93.4} \\ \midrule
		DenseASPP \cite{yang2018denseaspp} & 94.3 & 90.9 & \textbf{95.8} & 93.3 & 87.4 \\
		HA-DenseASPP & \textbf{96.6} & \textbf{96.8} & 95.0 & \textbf{95.9} & \textbf{92.1} \\ \midrule
		DUpsampling \cite{tian2019decoders} & 93.3 & 89.0 & \textbf{95.3} & 92.0 & 85.3 \\
		HA-DUpsampling & \textbf{95.9} & \textbf{96.0} & 94.2 & \textbf{95.1} & \textbf{90.6} \\ \midrule
		GSCNN \cite{takikawa2019gated} & 93.8 & 90.8 & \textbf{94.7} & 92.7 & 86.4 \\
		HA-GSCNN & \textbf{96.4} & \textbf{97.7} & 93.8 & \textbf{95.7} & \textbf{91.8} \\
		\bottomrule
	\end{tabular}
\end{table*}

\subsection{Experiment Setup}

In our experiments, six state-of-the-art networks: SegNet \cite{badrinarayanan2017segnet}, U-Net \cite{ronneberger2015u}, DeepLabv3+ \cite{chen2018encoder}, DenseASPP \cite{yang2018denseaspp}, DUpsampling \cite{tian2019decoders}, and GSCNN \cite{takikawa2019gated} are trained to validate the effectiveness and robustness of our proposed DS-Generator.  The networks trained on the augmented training sets are named as  ``HA-Network'', such as HA-U-Net and HA-DeepLabv3+. Furthermore, five metrics: a) accuracy, b) precision, c) recall, d) F-score and e) the intersection over union (IoU) are used to quantify the performance of the trained DCNNs.

Additionally, other conventional training data augmentation methods, such as translation and rotation, are also used in our experiments. The stochastic gradient descent with momentum (SGDM) optimizer is utilized to minimize the loss function, and the initial learning rate is set to $0.001$. Furthermore, we adopt the early-stopping mechanism \cite{goodfellow2016deep} on the validation set to reduce over-fitting problem. The DCNN performance is then quantified on the testing set, as presented in subsection \ref{sec.performan_evaluation}. Moreover, we select the best-performing model and fine-tune it for the result submission to the KITTI road benchmark \cite{Fritsch2013ITSC}.

\subsection{Performance Evaluation}
\label{sec.performan_evaluation}
This subsection evaluates the performance of our proposed DS-Generator both qualitatively and quantitatively. Examples of the experimental results on the KITTI \cite{Fritsch2013ITSC}, SYNTHIA \cite{HernandezBMVC17} and our SYN-Stereo road datasets are shown in Figs. \ref{fig.kitti_valid},  \ref{fig.synthia} and \ref{fig.carla}, respectively. We can clearly observe that the DCNNs trained on the augmented training set generally perform better than the same DCNNs trained on the original training set. The corresponding quantitative comparisons are given in Tables \ref{tab.kitti}, \ref{tab.synthia} and \ref{tab.carla}, respectively, where it can be seen that the F-score and IoU of the DCNNs trained on the augmented training set obtained by our proposed DS-Generator are improved by around 1.5-5.0\% and 2.8-7.3\%, respectively. Furthermore, HA-DeepLabv3+ performs better than all other DCNNs. Our analysis shows that, compared to the common training set augmentation operations, our proposed DS-Generator can leverage the relationship between multi-view images to perform more effective training data augmentation, and thus, benefit all state-of-the-art DCNNs for collision-free space detection.

\begin{figure*}[!t]
	\begin{center}
		\centering
		\includegraphics[width=0.99\textwidth]{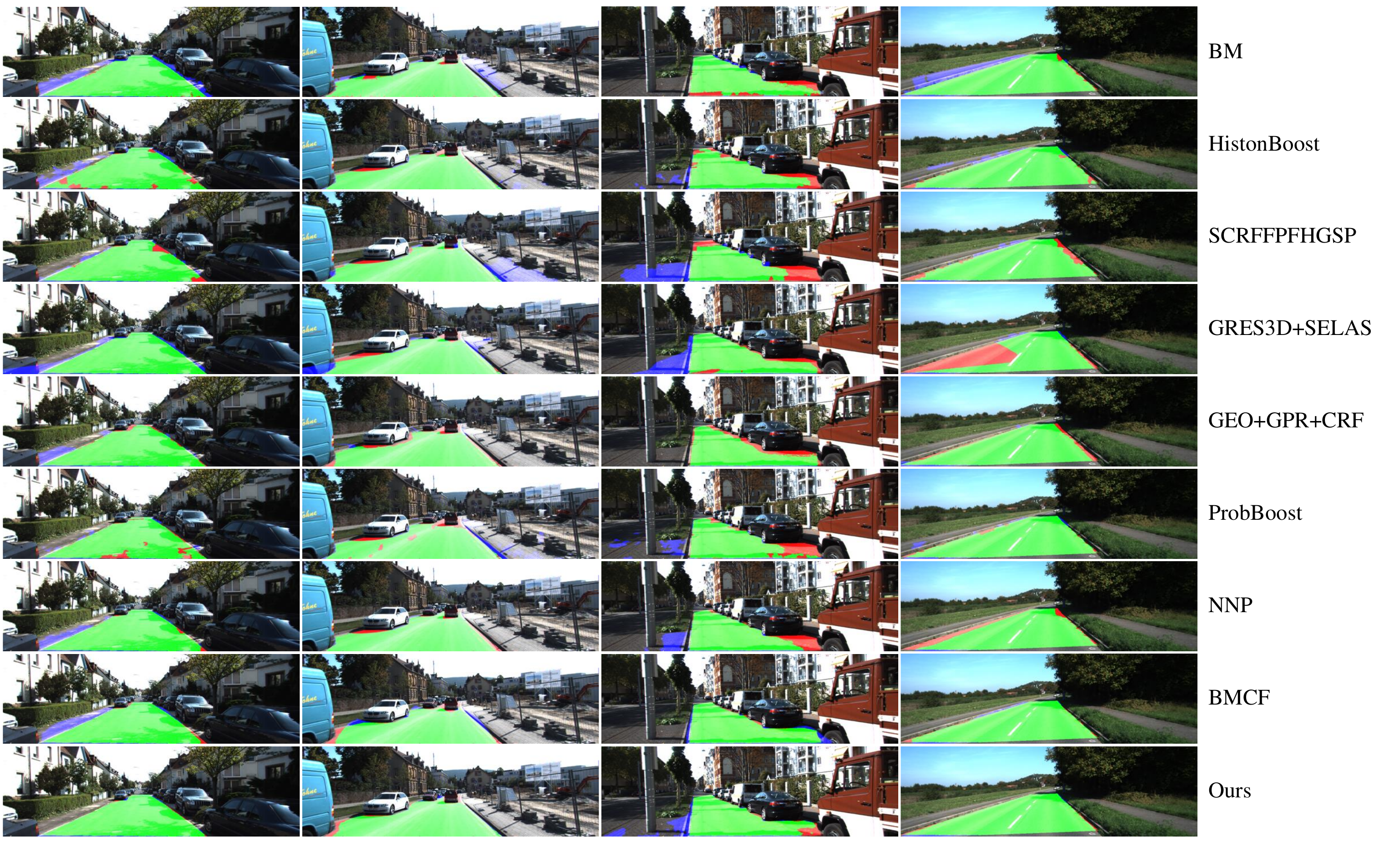}
		\centering
		\caption{Examples of the experimental results on the KITTI road benchmark,
			where the true positive, false negative and false positive pixels are shown in green, red and blue, respectively.}
		
		\label{fig.kitti_test}
	\end{center}
\end{figure*}
\begin{table*}[t!]
	\caption{Comparisons of the \textbf{Stereo Vision-Based} Collision-Free Space Detection Methods on the KITTI Road Benchmark, where $\uparrow$ Means Higher values are Better and $\downarrow$ Means Lower values are Better. Best Results are shown in bold type.}
	\centering
	\begin{tabular}{L{3.3cm}C{1.7cm}C{1.5cm}C{1.5cm}C{1.5cm}C{1.5cm}C{1.5cm}C{1.9cm}}
		\toprule
		Approach & MaxF ($\%$) $\uparrow$ & AP ($\%$) $\uparrow$ & PRE ($\%$) $\uparrow$ & REC ($\%$) $\uparrow$ & FPR ($\%$) $\downarrow$ & FNR ($\%$) $\downarrow$ & Runtime (s) $\downarrow$ \\ \midrule
		%		ANN \cite{vitor20132d} & 67.70 & 52.50 & 54.19 & 90.17 & 41.98 & 9.83 & 3 \\
		%		SPlane+BL \cite{einecke2014block} & 79.63 & 83.90 & 72.59 & 88.17 & 18.34 & 11.83 & 2 \\
		%		RES3D-Stereo \cite{shinzato2014road} & 81.08 & 81.68 & 78.14 & 84.24 & 12.98 & 15.76 & 0.70 \\
		BM \cite{wang2014color} & 83.47 & 72.23 & 75.90 & 92.72 & 16.22 & 7.28 & 2 \\
		HistonBoost \cite{vitor2014comprehensive} & 83.92 & 73.75 & 82.24 & 85.66 & 10.19 & 14.34 & 150 \\
		SCRFFPFHGSP \cite{gheorghe2015semantic} & 84.93 & 76.31 & 85.37 & 84.49 & 7.98 & 15.51 & 5 \\
		GRES3D+SELAS \cite{shinzato2015estimation} & 85.09 & 86.86 & 82.27 & 88.10 & 10.46 & 11.90 & 0.11 \\
		GEO+GPR+CRF \cite{xiao2017gaussian} & 85.56 & 74.21 & 82.81 & 88.50 & 10.12 & 11.50 & 30 \\
		ProbBoost \cite{vitor2014probabilistic} & 87.78 & 77.30 & 86.59 & 89.01 & 7.60 & 10.99 & 150 \\
		NNP \cite{chen20153d} & 89.68 & 86.50 & 89.67 & 89.68 & 5.69 & 10.32 & 5 \\
		BMCF \cite{wang2016multi} & 89.75 & 84.15 & 89.02 & 90.49 & 6.15 & 9.51 & 2.50 \\ \midrule
		HA-DeepLabv3+ (Ours) & \textbf{94.83} & \textbf{93.24} & \textbf{94.77} & \textbf{94.89} & \textbf{2.88} & \textbf{5.11} & \textbf{0.06} \\ \bottomrule
	\end{tabular}
	\label{tab.road}
\end{table*}

As mentioned above, we fine-tune our best-performing method, HA-DeepLabv3+\footnote{\url{www.cvlibs.net/datasets/kitti/eval_road_detail.php?result=4d39ae0a09df67b61c037ad3829f1a2c2b848f07}}, and submit its results to the KITTI road benchmark \cite{Fritsch2013ITSC}. Then, we compare our HA-DeepLabv3+ with eight state-of-the-art stereo vision-based collision-free space detection methods: 
%ANN \cite{vitor20132d}, SPlane+BL \cite{einecke2014block}, RES3D-Stereo \cite{shinzato2014road}, 
BM \cite{wang2014color}, HistonBoost \cite{vitor2014comprehensive}, SCRFFPFHGSP \cite{gheorghe2015semantic},
GRES3D+SELAS \cite{shinzato2015estimation}, GEO+GPR+CRF \cite{xiao2017gaussian}, ProbBoost \cite{vitor2014probabilistic}, NNP \cite{chen20153d}, and BMCF \cite{wang2016multi}, 
published on the KITTI road benchmark. Examples of the experimental results are shown in Fig. \ref{fig.kitti_test}. 
The quantitative comparisons are given in Table \ref{tab.road}. Readers can see that our HA-DeepLabv3+ is the best stereo vision-based collision-free space detection method, which achieves the highest MaxF (maximum F-score), AP (average precision), PRE (precision), REC (recall), FPR (false positive rate) and FNR (false negative rate). Furthermore, our method runs in real time and it is much faster than all other compared methods.

\section{Conclusion}
\label{sec.conclusion_future_work}
This paper proposed a novel training data augmentation approach, referred to as DS-Generator. It can generate additional driving scene images from multi-view vision data, such as stereo image pairs. Furthermore, we published a synthetic collision-free space detection dataset, named SYN-Stereo road dataset for research purposes. Extensive experimental results conducted with six state-of-the-art DCNNs on three datasets demonstrated the effectiveness of our DS-Generator, where the F-score and IoU of the DCNNs are improved by around 1.5-5.0\% and 2.8-7.3\%, respectively. Furthermore, HA-DeepLabv3+, our best-performing implementation, achieves the best overall performance compared to other stereo vision-based collision-free space detection algorithms published on the KITTI road benchmark.

\bibliographystyle{IEEEtran}

\end{document}